\documentclass[11pt]{article}

\usepackage[final]{acl}
\usepackage{amsmath}
\usepackage{times}
\usepackage{latexsym}
\usepackage[T1]{fontenc}

\usepackage[utf8]{inputenc}
\usepackage{tablefootnote}
\usepackage{microtype}
\usepackage{booktabs}
\usepackage{inconsolata}
\usepackage{graphicx}
\usepackage{amssymb}
\usepackage{pifont}

\newcommand{\xmark}{\ding{55}}%

\title{Encoder Fine-tuning with Stochastic Sampling Outperforms Open-weight GPT in Astronomy Knowledge Extraction}

\author{Shivam Rawat \hspace{3mm} \textbf{Lucie Flek} \hspace{3mm} \textbf{Akbar Karimi} \\ 
    Bonn-Aachen International Center for Information Technology,  University of Bonn, Germany \\
    Lamarr Institute for Machine Learning and Artificial Intelligence, Germany\\
    \texttt{s.rawat@uni-bonn.de}}

\begin{document}
\maketitle
\begin{abstract}
Scientific literature in astronomy is rapidly expanding, making it increasingly important to automate the extraction of key entities and contextual information from research papers. In this paper, we present an encoder-based system for extracting knowledge from astronomy articles. Our objective is to develop models capable of classifying telescope references, detecting auxiliary semantic attributes, and recognizing instrument mentions from textual content. To this end, we implement a multi-task transformer-based system built upon the SciBERT model and fine-tuned for astronomy corpora classification. To carry out the fine-tuning, we stochastically sample segments from the training data and use majority voting over the test segments at inference time. Our system, despite its simplicity and low-cost implementation, significantly outperforms the open-weight GPT baseline. 
\end{abstract}

\section{Introduction}

Evaluating the scientific influence of an astronomical observatory often relies on quantitatively reviewing publications that use its data, typically by constructing bibliographies that link datasets to scholarly articles \cite{kurtz2000nasa, accomazzi2011linking, henneken2011linking, grezes2023function}. This process enables bibliometric analyses and supports scientific reproducibility, although it remains labor-intensive and depends heavily on expert knowledge. 
While some tools for literature curation offer inexpensive solutions by relying on keyword matching  \cite{dai2022detecting}, others have used recent generative transformer-based models \cite{vaswani2017attention, feng2025reliability}. Their self-attention mechanism enables the modeling of long-range dependencies in text, and their ability to generate and classify human-like language has led to successful cross-domain applications \cite{chae2023large, aly2025evaluation}.
However, while LLMs offer some advantages in accurately extracting general and fine-grained information from domain-specific astrophysical texts \cite{shao2024astronomical},
they are computationally expensive to deploy and are not always optimized for specialized scientific concepts. As a result, a lightweight, domain-adapted method that can support large-scale curation without prohibitive resource costs is needed.
In this work, we present a simple, low-cost approach for classifying and inferring instrumentation information from astrophysical literature. We show that it significantly outperforms the 20B-parameter LLM baseline\footnote{\url{https://huggingface.co/openai/gpt-oss-20b}} on this task, demonstrating the value of domain alignment over sheer model size. Our contributions are twofold: (1) we implement an efficient model that can be deployed at scale; and (2) we provide empirical evidence that lightweight, domain-specific methods can surpass much larger general-purpose LLMs.
By enabling accurate and scalable linkage between observational data and the scholarly record, our approach supports both bibliometric evaluation and scientific reproducibility and highlights the importance of tailored NLP solutions for scientific domains.

\section{Task Description}

The Telescope Reference and Astronomy Categorization Shared Task (TRACS) at IJCNLP-AACL 2025 \cite{grezes-etal-2025-tracs} presents us with a unique opportunity to apply natural language processing techniques to astrophysical literature and derive actionable insights to assist the scientific method.

\begin{figure*}
    \centering
    \includegraphics[scale=0.37]{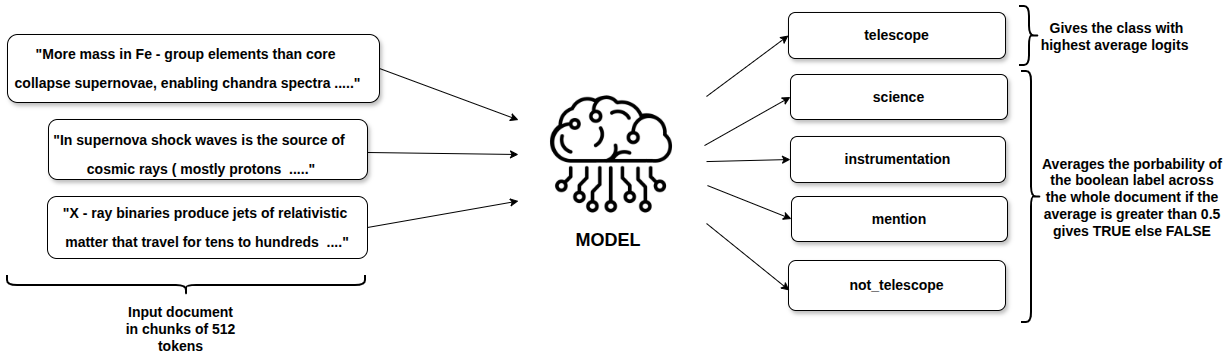}
    \caption{System design for the shared task. Input documents are chunked into equal segments with 512 tokens. Each segment is given the same label as the original input and is used to fine-tune the model. At inference time, we use majority voting to assign the test labels.}
    \label{fig:model}
\end{figure*}

\subsection{Objective}

The objective of the given task is that given an astrophysical text, we need to train a language model that can infer the information about the telescope instrument being used. The model should be able to identify the telescope being used in the text and also in what capacity it is being used. To quantify it, the text needs to be classified into 4 boolean labels, which are "science", "instrumentation", "mention", and "not\_telescope".

\subsection{Dataset}

The dataset provided for the TRACS@WASP task is full papers or fragments of papers that are taken from SciX\footnote{\url{https://www.scixplorer.org}} and are meticulously annotated by the domain experts.
The dataset is provided in a CSV format. Each row consists of the following elements:
\begin{itemize}
    \item "bibcode": A unique string for entry identification in the SciX database, which is necessary for organization and traceability.
    \item "telescope": The name of the telescope, which is referenced in the entry.
    \item "author", "year": The metadata on the researchers and the time of publication of the entry.
    \item "title", "abstract", "body", "acknowledgments", "grants": The textual content of the entry, which are essentially different parts of the research document, is split according to these labels.
    \item "science", "instrumentation", "mention", "not\_telescope": These are boolean labels which classify the entry according to how the papers use the data from the telescopes.   
\end{itemize}
For the training dataset, the annotated labels that the model needs to train and predict are the multiclass label "telescope" and the four boolean labels "science", "instrumentation", "mention", and  "not\_telescope". The data for training, as one can infer, is the textual information for the research paper split into "title", "abstract", "body", "acknowledgments", and "grants".

\subsection{Data Statistics and Preprocessing}\label{sec:dataset}

Diving into the statistics of the provided dataset, it consists of 80385 unique entries spanning 4 decades for three telescopes. These are the Hubble Space Telescope (HST), the Chandra X-ray Observatory (CXO), and the James Webb Space Telescope (JWST). Also, among the four boolean labels, "science" and "mention" are fairly evenly distributed, but the remaining two are quite skewed, with the majority of entries being the boolean label "FALSE".
For the full entry text to be processed by our model, we convert the dataset into multiple JSON files. First, we concatenate the content of the fields "title", "abstract", "body", "acknowledgments", and "grants". Then, we split this string into chunks of 512 tokens, which were then saved in the JSON format along with the labels. Each JSON file contains 1000 entries, which are chunked in the manner described. Finally, for 80385 rows in the CSV file, we get 81 JSON files, which are then used for training purposes (Figure \ref{fig:model}). These preprocessed JSON files are used as training data.

\begin{table}
    \centering
    \small
    \setlength{\tabcolsep}{2pt}
    \begin{tabular}{l|c|c|c|c|c}
    Model& Optimizer & LR & Scheduler& Batch Size&Epochs\\
    \hline
     SciBERT  & AdamW& 2e-5& linear&8&4\\
    \end{tabular}
    \caption{Model training hyperparameters}
    \label{tab:config}
\end{table}

\section{Experiments and Results}

\subsection{Model Selection}
To perform the task, we opted for the SciBERT model \cite{beltagy2019scibert} since it is a pretrained language model designed to enhance natural language understanding within the scientific domain. Built upon the foundational BERT architecture \cite{devlin2019bert}, SciBERT extends its capabilities by being trained on a large corpus of scientific publications sourced from the Semantic Scholar database \cite{ammar2018construction}. This domain-specific pretraining enables SciBERT to capture the specialized vocabulary, structure, and linguistic patterns prevalent in scientific writing, which are often underrepresented in general-domain corpora.

The model maintains the same architecture as BERT-Base but introduces a newly constructed vocabulary, SciVocab, tailored to the scientific domain. This vocabulary shares only about 42\% overlap with BERT’s original WordPiece vocabulary, highlighting the substantial linguistic differences between general and scientific texts \cite{beltagy2019scibert}. Through this adaptation, SciBERT demonstrates superior performance across a range of scientific NLP tasks, including named entity recognition, relation classification, sentence classification, and dependency parsing, outperforming general-domain models on domain-specific benchmarks. Its advantages are particularly pronounced in biomedical and life science applications, where scientific terminology and context play crucial roles in comprehension and information extraction.

\subsection{Experiments and Results}
We report the results for two sets of experiments that showed a marginal difference in their performance. Both these experiments achieved the 6\textsuperscript{th} rank in the competition leaderboard.

In our approach, we initially do a baseline run to measure the scope of improvement. The pretrained SciBERT encoder was used without any fine-tuning, while the classification heads remained randomly initialized. The [CLS] token representations from each chunk were processed by the random classification heads to generate logits for both telescope and boolean labels. Predictions were then aggregated across all chunks to produce final outputs. We call this experiment SciBERT\_v1.

Following this, we begin our training procedure. In the first experiment, we use the first 512 tokens from each entry, along with the entry-level classification labels. This results in a dataset comprising approximately 41 million tokens. The training hyperparameters used are listed in Table~\ref{tab:config}. The loss function governing this training process is the sum of the cross-entropy loss for the multiclass label (e.g., the “telescope” label) and the BCEWithLogits loss for the four boolean labels.

\begin{table}
    \centering
    \begin{tabular}{c|c}
    \toprule
       Model  & Macro F1 score \\
       \midrule
        SciBERT\_base & 0.18\\
        Random baseline & 0.24 \\
        Openai-gpt-oss-20b\tablefootnote{\url{https://ui.adsabs.harvard.edu/WIESP/2025/shared_task}} & 0.31 \\
        \textit{SciBERT\_v1} & \textit{0.72} \\
        \textbf{SciBERT\_v2}& \textbf{0.73} \\
        \bottomrule
    \end{tabular}
    \caption{Performance metrics. Here, '\_base' represents the baseline run, '\_v1' is the SciBERT model trained with the initial 512 tokens, and '\_v2' is the SciBERT model trained on the 10 random chunks from each entry.}
    \label{tab:result}
\end{table}

Next, we carry out a similar experiment, but instead of using just the first 512 tokens, we use 10 random chunks from each entry (if an entry has fewer than 10 chunks, we consider all of them). We call this experiment SciBERT\_v2. All the resulting chunks get labeled the same as the full entry itself. This was done to give a fair chance to the other chunks of the same entry to contribute to the training part, specifically the acknowledgment and grants, which often contain direct references to instrumentation.  For this experiment, the dataset comprises approximately 410 million tokens. The remaining part of the training process (the hyperparameters, loss function, etc) was similar to the previous experiment.

\begin{table*}
    \centering
    \small
    \begin{tabular}{l|c|c|c|c|c|c|c|c|c|c}
         Paper ID & \multicolumn{2}{|c}{Telescope} & \multicolumn{2}{|c}{Science} & \multicolumn{2}{|c}{Instrument} & \multicolumn{2}{|c}{Mention} & \multicolumn{2}{|c}{Not\_telescope} \\
         \toprule
         &GT&Pred&GT&Pred&GT&Pred&GT&Pred&GT&Pred\\
         \midrule
         2014H...6C\_CHANDRA& CHANDRA & CHANDRA \checkmark &1&1 \checkmark&0&0 \checkmark&0&0 \checkmark&0&0 \checkmark \\
         2001t...7M\_CHANDRA& CHANDRA & CHANDRA \checkmark &0&0 \checkmark&0&0 \checkmark&1&1 \checkmark&0&0 \checkmark \\
         2008l...8S\_HST& HST & HST \checkmark &1&1 \checkmark&0&0 \checkmark&0&0 \checkmark&0&0 \checkmark\\
         2012A...4S\_CHANDRA& CHANDRA & CHANDRA \checkmark &0&0 \checkmark&0&0 \checkmark&1&1 \checkmark&0&0 \checkmark \\
         \midrule
         2011A...1M\_CHANDRA& CHANDRA & CHANDRA \checkmark &1&0 \textcolor{red}{\xmark} &0&0 \checkmark&0&1 \textcolor{red}{\xmark} &0&0 \checkmark \\
         2020S...9M\_HST& HST & JWST \textcolor{red}{\xmark} &0&0  \checkmark&0&0 \checkmark&0&1 \textcolor{red}{\xmark} &1&0 \textcolor{red}{\xmark} \\
         2022s...1W\_CHANDRA& CHANDRA & CHANDRA \checkmark &1&0  \textcolor{red}{\xmark}&0&0 \checkmark&0&1 \textcolor{red}{\xmark} &0&0 \checkmark \\
         2000H...7S\_CHANDRA& CHANDRA & CHANDRA \checkmark &1&0  \textcolor{red}{\xmark}&0&0 \checkmark&0&1 \textcolor{red}{\xmark} &0&0 \checkmark \\
         \bottomrule
    \end{tabular}
    \caption{Model prediction examples}
    \label{tab:eg}
\end{table*}

These models were tested on the test dataset, which consisted of 9194 entries. These were also preprocessed in the same way as the training dataset. To quantify the model's classification capability appropriate metric is needed. For classification tasks where there is label imbalance F1 score is most widely used. The F1 score provides a balanced measure of a model’s precision and recall, which is especially important for imbalanced datasets, which we have as we discussed in the \ref{sec:dataset} already. Now, since we have 5 classes to predict, we will have an F1 score per class. So we consider the macro F1 score as the model performance metric given as:
\begin{align}
\text{Model}_{F1} &= \frac{\text{multiclass}_{F1}+\frac{1}{N}\sum_{i}\text{bool}_{F1,i}}{2}
\end{align}

\noindent
where multiclass is for the "telescope" label and bool for the four boolean classes "science", "instrumentation", "mention", and "not\_telescope". The metrics of the trained model are compared to the baseline in Table~\ref{tab:result}. As we can see, the results from our two experiments are similar ($0.72$ and $0.73$). However, they significantly outperform the LLM baseline, which has a performance of $0.31$, as well as our own baseline, which is the same model without fine-tuning ($0.18$). This can be attributed to the domain-specific fine-tuning, which allowed our trained models to be specialized classifiers.

\section{Error Analysis}

To gain deeper insights into the limitations of our approach, we perform an error analysis. Since ground-truth labels for the test set are not available, this analysis is conducted on the validation split of the training data, which was also used for evaluation during model development.

\begin{figure}[t]
    \includegraphics[scale=0.45]{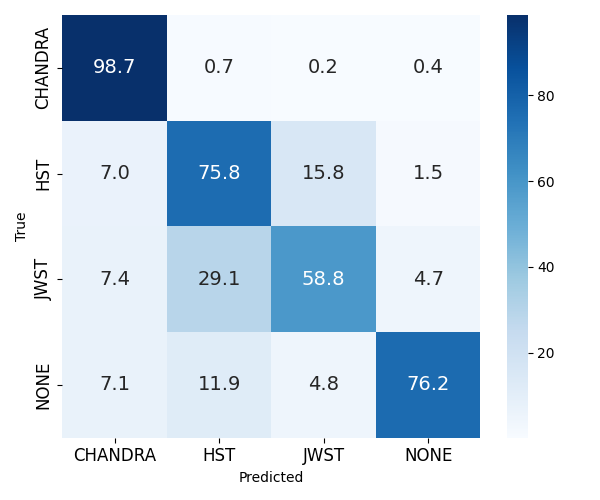}
    \caption{Telescope predictions confusion matrix}
    \label{fig:telescope}
\end{figure}

In Table \ref{tab:eg}, we present selected example predictions from our best-performing model, SciBERT\_v2. In the misclassified cases, we observe that the boolean labels “science” and “mention” tend to be mispredicted more frequently. This behavior is likely because these labels are highly context-dependent, requiring a nuanced understanding of the surrounding textual semantics. In contrast, the labels “instrument” and “not\_telescope” are generally easier to predict correctly, as their identification primarily depends on the explicit mention of instrument names rather than broader contextual cues. This issue could be alleviated by employing models capable of handling longer context windows or those pretrained on domain-specific astronomical corpora.
Furthermore, for the telescope classification, a clear trend emerges (Figure \ref{fig:telescope}): the model achieves the highest accuracy for CHANDRA, followed by HST, and then JWST. Also, we see increased false predictions, i.e., more confusion for the HST and JWST classes. The reason behind this could be the naming scheme of the classes for this label. The classes of the Hubble Space Telescope (HST) and the James Webb Space Telescope (JWST) share the words "Space" and "Telescope" that might have confused the model predictions, while CHANDRA is more distinct.

\section{Conclusion and Future Work}

We introduced our system for the telescope reference and astronomy categorization. Leveraging the SciBERT model, our method utilizes domain-adapted language representations to automatically identify telescope mentions and their contextual roles within astrophysical literature. 
We showed that fine-tuning SciBERT on random segments selected from the article data considerably improves model performance and significantly outperforms the LLM baseline.
Looking ahead, we aim to further enhance the framework by exploring transformers with extended context windows and models pretrained on astronomy-specific corpora, which could help capture the nuanced contextual cues required for labels such as science and mention. We also plan to investigate data balancing strategies and contrastive learning methods to mitigate class skewness in telescope categories and improve robustness across less frequent instruments.

\section{Limitations}

The limitations of this work primarily stem from the inherent challenges of modeling complex scientific text and the class imbalance in the dataset. Although our framework effectively captures domain-specific semantics, the context-dependent nature of certain labels makes it prone to misclassification, suggesting that the current model’s context window may be insufficient to fully capture subtle relationships between telescope usage and scientific context.
Furthermore, the reliance on weakly supervised labels may introduce annotation noise, affecting the precision of the boolean attribute detection. The telescope classification results also reflect dataset skewness, where classes such as CHANDRA are overrepresented, leading to uneven performances across telescope types. 

\section*{Acknowledgments}

This work was supported by the BMFTR and the state of North Rhine-Westphalia as part of the Lamarr Institute for Machine Learning and Artificial Intelligence, the Transdisciplinary Research Area (TRA) Matter at the University of Bonn, and the AISafety Project, funded under the BMFTR grant proposal No. 05D23PD1.

\bibliography{custom}

\end{document}